%% file: main.tex
\pdfoutput=1

\documentclass[11pt]{article}

\usepackage[]{acl}

\usepackage{times}
\usepackage{latexsym}

\usepackage[T1]{fontenc}

\usepackage[utf8]{inputenc}
\usepackage{float}
\usepackage{microtype}
\usepackage{hyperref}
\usepackage{url}
\usepackage{graphicx}
\usepackage{booktabs}
\usepackage{multirow}
\usepackage{times}
\usepackage{latexsym}
\usepackage{algorithm}
\usepackage{algorithmic}
\usepackage{inconsolata}
\usepackage{amsmath}
\usepackage{amssymb}
\usepackage{mathtools}
\usepackage{amsthm}
\usepackage{tabularx}
\usepackage{enumitem}
\usepackage{multirow}
\usepackage{multicol}
\usepackage{makecell}
\usepackage{booktabs}
\usepackage{subfigure}
\title{Demonstration Augmentation for Zero-shot In-context Learning}

\author{Yi Su$^{1}$\thanks{\; Equal Contribution.}, Yunpeng Tai$^1$\footnotemark[1], Yixin Ji$^1$, Juntao Li$^{1}$, Bowen Yan$^2$, Min Zhang$^{1}$ \\  
 $^{1}$School of Computer Science and Technology, Soochow University\\
  $^{2}$Department of Computer Science and Technology, Tsinghua University\\
 \texttt{yisunlp@outlook.com};  \texttt{yunpengtai.typ@gmail.com}\\
 \texttt{yanbw@mail.tsinghua.edu.cn};\\
  \texttt{\{ljt,minzhang\}@suda.edu.cn} \\
 }

\begin{document}
\maketitle
\input{paper/abs}
\input{paper/intro}
\input{paper/preliminary}
\input{paper/method}
\input{paper/exp}

\input{paper/exp_res}
\input{paper/analysis}
\input{paper/related}
\input{paper/conclusion}
\section*{Limitations}
While DAIL has demonstrated superior accuracy and inference speed compared to all baselines, it is important to acknowledge its limitations:

\begin{itemize}[leftmargin=*]
    \setlength{\itemsep}{0pt}
    \setlength{\parskip}{0pt}
    \item Obtaining the Entropy Score in DAIL requires accessing logits, which can be challenging when using APIs for inference. This introduces deployment challenges for DAIL in real-world applications.
    
    \item Our validation of DAIL's performance has primarily focused on MMLU and BBH, both of which involve multiple-choice tasks. Its effectiveness in open-domain text generation tasks has yet to be confirmed.
    
    \item Storing previously inferred samples poses potential privacy concerns and increases the risk of privacy breaches. In scenarios prioritizing data security, DAIL may not be the most suitable solution.
\end{itemize}

\section*{Acknowledgements}
We want to thank all the anonymous reviewers for their valuable comments. Juntao Li and Bowen Yan are the corresponding authors. This work was supported by the National Science Foundation of China (NSFC No. 62206194), the Natural Science Foundation of Jiangsu Province, China (Grant No. BK20220488), Young Elite Scientists Sponsorship Program by CAST (2023QNRC001), and Supercomputing Center in Yancheng, Grant No. 20231001.

\bibliography{anthology,custom}
\clearpage
\appendix
\input{paper/appendix}

\end{document}

%% file: paper/abs.tex
\begin{abstract}
Large Language Models (LLMs) have demonstrated an impressive capability known as In-context Learning (ICL), which enables them to acquire knowledge from textual demonstrations without the need for parameter updates.
However, many studies have highlighted that the model's performance is sensitive to the choice of demonstrations, presenting a significant challenge for practical applications where we lack prior knowledge of user queries.
Consequently, we need to construct an extensive demonstration pool and incorporate external databases to assist the model, leading to considerable time and financial costs.
In light of this, some recent research has shifted focus towards zero-shot ICL, aiming to reduce the model's reliance on external information by leveraging their inherent generative capabilities. 
Despite the effectiveness of these approaches, the content generated by the model may be unreliable, and the generation process is time-consuming.
To address these issues, we propose \textbf{D}emonstration \textbf{A}ugmentation for \textbf{I}n-context \textbf{L}earning (\textbf{DAIL}), which employs the model's previously predicted historical samples as demonstrations for subsequent ones.
DAIL brings no additional inference cost and does not rely on the model's generative capabilities.
Our experiments reveal that DAIL can significantly improve the model's performance over direct zero-shot inference and can even outperform few-shot ICL without any external information. Our code is available at \url{https://github.com/yisunlp/DAIL}.
\end{abstract}

%% file: paper/intro.tex
\section{Introduction}
Large Language models (LLMs) have recently gained widespread attention due to their numerous advantages, including user-friendly interactions, convenient applications, and zero-shot capabilities \citep{wei2021finetuned,chatgpt,scao2022bloom,zhang2022opt,OpenAI2023GPT4TR,touvron2023llama,baichuan2023baichuan2}. 
However, the expanding parameter scale of LLMs poses a significant challenge to fine-tuning, demanding considerable investments in both time and computational resources.
Therefore, In-context Learning (ICL), a method enabling LLMs to acquire knowledge through textual demonstrations without the need for parameter updates, has become increasingly important in recent times \citep{wei2021finetuned,dong2022survey}.

Conditioning on some input-label pairs (demonstrations), LLMs can rapidly acquire the ability to solve new tasks in a few-shot manner just by combining the demonstrations and the sample together\citep{radford2019language,brown2020language}.
However, many studies indicate that the model's performance is sensitive to the choice of demonstrations \citep{zhang2022active,liu2022makes,hao2022structured,lu2022fantastically}.
In extreme cases, inadequately chosen demonstrations can significantly degrade the model's performance, causing a drastic drop from State-of-the-Art to near-random \citep{liu2022makes}.  
To address this challenge, researchers have proposed various solutions, including demonstration selection, calibration, and arrangement \citep{rubin2022learning,min2022rethinking,min2022metaicl,zhao2021calibrate,chen2022improving,yoo2022ground,min2022noisy}. 
These methods can improve the model's performance and stability under ICL across many tasks.

Nevertheless, these approaches are still insufficient to ensure the reliable application of ICL in real-world scenarios, where our prior knowledge of user queries is often limited.
Consequently, it is a common practice to construct an extensive demonstration pool, harness external databases, and implement various strategies such as selection, calibration, and arrangement methods, as mentioned earlier, to deal with all kinds of queries from users.
However, this process entails a significant investment in time and financial resources.
Therefore, some researchers \citep{zhang2022automatic,kim2022self,lyu2023z,chen2023self} attempt to alleviate the reliance on external information by proposing zero-shot ICL.
These approaches leverage the model's generative capabilities to produce the required information for the inference process.
In this context, zero-shot ICL offers a promising avenue for more efficient and cost-effective deployment for ICL.

These methods can reduce the dependence on external information, but the quality of the content generated by the model cannot be guaranteed, which may pose some potential risks.
Furthermore, the generation process is time-consuming, which can bring additional costs during inference.
To address these problems, we propose \textbf{D}emonstration \textbf{A}ugmentation for \textbf{I}n-context \textbf{L}earning (\textbf{DAIL}), which employs the model's previously predicted historical samples as demonstrations for subsequent ones\footnote{At the beginning, we use zero-shot inference because there is no historical samples.}.
Specifically, we only need to maintain a memory bank of a small size $M$ and define the entry, selection, and deletion strategies.
During the inference phase, the selection strategy chooses the most suitable demonstrations from the memory bank.
Subsequently, we use the entry strategy to add the predicted sample to the memory bank.
Upon reaching maximum capacity, we use the deletion strategy to remove some stored samples.
Our experiments on different benchmarks and models demonstrate the effectiveness of DAIL.

Overall, our contributions in this work include:
\begin{itemize}[leftmargin=*]
\setlength{\itemsep}{0pt}
\setlength{\parskip}{0pt}
\item We point out the potential limitations of previous zero-shot methods in stability and inference time.
\item We introduce DAIL, an easy yet effective method to enhance zero-shot ICL.
\item Our experiments reveal that DAIL can significantly improve the model's performance over direct zero-shot inference and can even outperform few-shot ICL without any external information.
\end{itemize}

%% file: paper/preliminary.tex
\section{Preliminary}
\subsection{Problem Formulation}
In this subsection, we briefly summarize the inference process of ICL.
A Large Language Model (LLM) can be formalized as a function $f:X \xrightarrow{} Y$, mapping the input space $X$ to the output space $Y$.
The corresponding dataset comprises a set of labeled demonstrations $\{x_s^i, y_s^i\}_{i=1}^{n_s}$ and a set of unlabeled queries $\{x_t^i\}_{i=1}^{n_t}$.
Then, a carefully crafted template $t$ is utilized to transform each sample into a natural language sentence that the model can process.
During the inference stage for a given query, $K$ demonstrations are selected from the demonstration pool based on a selection strategy such as TopK \citep{liu2022makes}. 
Subsequently, these chosen demonstrations and the query are combined to construct the input sequence for the model:
\begin{equation}
    Input = \{t(x_s^1,y_s^1),...,t(x_s^K,y_s^K),t(x_t^i)\},
\end{equation}
where $t(\cdot)$ is the transformation of the template.

The model processes the input sequence and generates the final output, denoted as:
\begin{equation}
    Output = V(f(Input)),
\end{equation}
where $V$ is a mapping function that converts the model's output into a label in the label space.
It can be a text-level matching function or a selection mechanism based on probability or perplexity.

\subsection{Potential Risks of Previous Methods}
\begin{table}[t]
\centering
\resizebox{\columnwidth}{!}{
\begin{tabular}{lcc}
\toprule
\textbf{Method} & \textbf{Inputs} & \textbf{Labels} \\
\hline 
AUTO-COT \citep{zhang2022automatic} & from training set & no need \\
Z-ICL \citep{lyu2023z} & from external corpus & no need \\
SG-ICL \citep{kim2022self} & no need & given \\
Self-ICL \citep{chen2023self} & no need & no need \\
DAIL (Ours) & no need & no need \\
\bottomrule
\end{tabular}}
\caption{\label{tab: resources} A comparison to prior attempts on zero-shot ICL. Self-ICL and DAIL do not require any external information to construct demonstrations.}
\end{table}
While previous works have delved into zero-shot ICL, they mainly focus on reducing the reliance on labeled demonstrations and are not entirely independent of external information (Table \ref{tab: resources} shows the resources needed for each method).
We concentrate on a setting that requires no external information, aiming to minimize the extra cost.
Moreover, these methods rely on the generative capabilities of the model, so there may be issues with poor generation quality and increased inference time.

\begin{figure}[t]
    \centering
    \includegraphics[width=0.8\hsize]{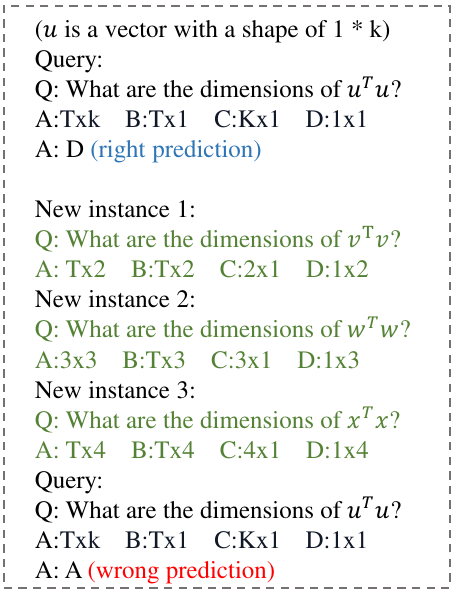}
    \caption{A bad case for Self-ICL, the quality of the generated samples is poor, with repeated options, false labels, and too similar semantics, which leads to the decline of the model's performance. For simplicity of the figure, we omit the generated labels of demonstrations.}
    \label{fig:preliminary}
\end{figure}

\begin{figure}[t]
    \centering
    \includegraphics[width=0.95\hsize]{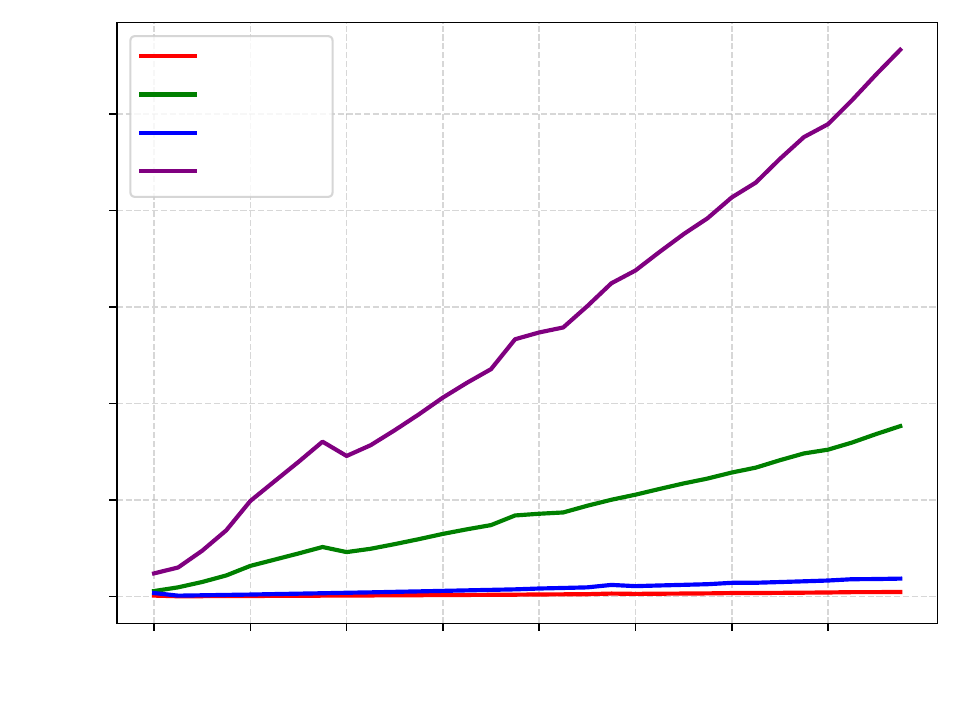}
    \caption{Time consumption (in seconds) for different methods and sequence lengths (batch size = 16). We use LLaMA-2-7B \citep{touvron2023llama} as the base model. Encode: cost of encoding n tokens. Generate: cost of generating n tokens. 3-shot: ICL with three demonstrations. For simplicity, we assume that all the demonstrations generated by the model have the same sequence length as the query.}
    \label{fig:speed}
\end{figure}

We take Self-ICL \citep{chen2023self} as an example.
Self-ICL performs the following three actions upon receiving a question: 1) It uses a human-designed prompt to guide the model in generating $K$ new, related, and diverse questions based on the original question. 2) It employs zero-shot inference to obtain the answers for the generated $K$ questions respectively. 3) It concatenates these $K$ questions and their answers to serve as demonstrations for ICL.
In our experiments, we find that Self-ICL relies heavily on the generative capability of the model. When the model generates poor demonstrations, it will hurt the performance of ICL (Figure \ref{fig:preliminary}).
Furthermore, Self-ICL requires more queries and token consumption than direct zero-shot inference, resulting in increased inference costs.
This is particularly pronounced in its generation process (Figure \ref{fig:speed}), where the expense of generating a token exceeds that of encoding a token. 
Consequently, this poses a challenge to computing resources during deployment for Self-ICL.

To address these challenges, we need to obtain more reliable demonstrations at a lower cost. 
It is intuitive that text provided by humans typically have higher quality than that generated by models. With the absence of external information, the human-supplied text available to LLM is only user queries.
Hence, our strategy involves leveraging previously predicted historical samples as demonstrations. This offers several advantages over other zero-shot ICL methods, including superior text quality, independence from the model's generative capabilities, and lower acquisition costs.

%% file: paper/method.tex
\section{Method}
\begin{figure*}[t]
    \centering
    \includegraphics[width=0.98\hsize]{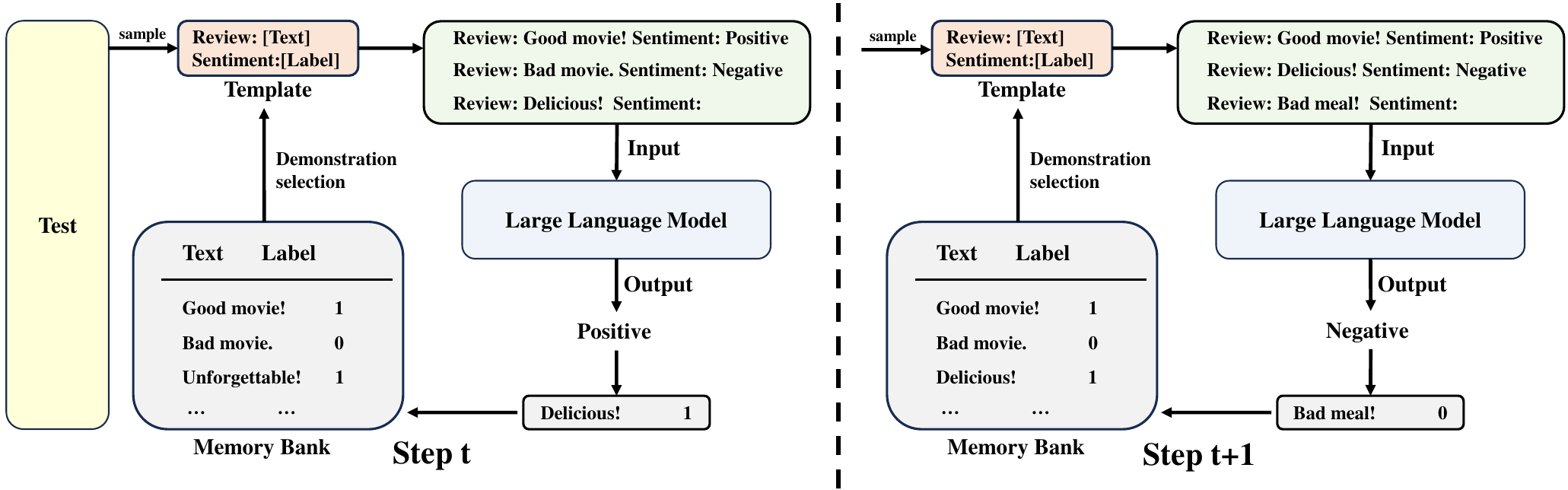}
    \caption{Overview of our method. After each inference, we combine the current query with the model's output and add them to the memory bank. After step t, the sample is added to the memory bank and then used as a demonstration at step t+1.}
    \label{fig:method}
\end{figure*}
Figure \ref{fig:method} illustrates an overview of our method.
Suppose we have a LLM $f$ and a set of user-issued queries to respond. We initialize a memory bank with a maximum capacity of $M$.
At step $0$, the memory bank is empty, and we directly process the query using zero-shot inference.
At step $t$, when a new query arrives, we employ the selection strategy to search for $K$ samples in the memory bank.
If the number of samples in the memory bank is less than $K$, we extract all demonstrations from the bank.
Following the model's output, we use the entry strategy to add the current sample, i.e., the current query paired with the correlated model response, to the memory bank.
At step $t+1$, the sample from the previous step $t$ has been in the memory bank and can be selected as a demonstration to help answer the new query.
In the following subsections, we will elaborate the details of our entry, selection, and deletion strategies, which manage the dynamic data flow within our memory bank.

\subsection{Entry Strategy}
Following the processing of each query, we combine the query and the corresponding response into a sample and directly add the sample to the memory bank. Despite its simplicity, the entry strategy is a fundamental building block in our method.
\subsection{Selection Strategy}
Our selection strategy involves assigning a score to each sample in the memory bank and selecting those with the highest scores.
Each sample's score comprises two factors: the Selection Score and the Entropy Score. 
Assuming the query that the model need to respond is $x$, and a sample in the memory bank is $\hat{x}$, we will explain how to compute the Selection Score and the Entropy Score for them.

\subsubsection{Selection Score}
The Selection Score primarily comes from existing demonstration selection methods.
We have experimented with various selection methods, including random selection, BM25 \citep{robertson2009probabilistic}, and TopK \citep{liu2022makes}.
\paragraph{Random score}
Under random selection, all samples are chosen with equal probability. Therefore, we assign a score of 0 to all samples.
\paragraph{BM25 score}
The BM25 algorithm \citep{robertson2009probabilistic} is a well-known information retrieval method based on the Okapi TF-IDF algorithm \citep{ramos2003using}.
It is widely employed for ranking queries in information retrieval tasks.
We can leverage it to compute a similarity score between $\hat{x}$ and $x$.
\begin{equation}
score_s=BM25(\hat{x},x),
\end{equation}
where $BM25(\cdot)$ represents the BM25 algorithm.
\paragraph{TopK score} Noting that selecting demonstrations with semantics closer to the query enhances the performance of ICL \citep{liu2022makes}, we utilize Sentence-BERT \citep{reimers2019sentence} to calculate the similarity between $\hat{x}$ and $x$.
\begin{equation}
    score_s= cos(emb(\hat{x}),emb(x)),
\end{equation}
where $emb(\cdot)$ denotes the process of computing the hidden states of a sentence with Sentence-BERT.

\subsubsection{Entropy Score}
Intuitively, samples with lower entropy should be prioritized for selection because this suggests that the sample is simpler and the pseudo-labels are more reliable \cite{su2023beware}. 
We can compute the entropy of a sample with the following formula: 
\begin{equation}
    score_e= - \sum_{c} p\left(y_c|x\right) \log p\left(y_c|x\right),
\end{equation}
where $p\left(y_c|x\right)$ is the next token prediction probability of $x$ provided by the model.
\subsubsection{Final Selection Strategy}
The final score consists of two components: the Selection score and the Entropy Score. We normalize each score to mitigate the differences between the aforementioned two types of scores.
\begin{equation}
    score= N\left(score_s\right) - \alpha * N\left(score_e\right),
\end{equation}
where $N(\cdot)$ stands for normalization, and $\alpha$ is a manually set hyper-parameter to balance the weight of the two scores.

In addition to the three selection methods above, we also utilize DPP \citep{kulesza2011k} to refine the TopK selections, aiming to enhance the diversity of the demonstrations.
Specifically, we employ the TopK Score and Entropy Score to select some candidates and apply the DPP algorithm to choose $K$ demonstrations from these candidates.
In total, we consider four selection methods, and their comparison will be discussed in Section \ref{select}.

\input{table/mmlu}

\subsection{Deletion Strategy}
Upon reaching full capacity, we delete half of the samples from the bank.
We explore three different deletion strategies: Random, FIFO, and Diverse.
\paragraph{Random} Randomly select half of the samples and delete them from the bank.
\paragraph{FIFO} Delete the samples that entered the bank earlier (First-In-First-Out).
\paragraph{Diverse} We aim to preserve the diversity of the samples in the bank after deletion.
We employ the TopK Score mentioned above to calculate the similarity between each sample and the entire bank.
Subsequently, we delete the samples with higher similarity, thereby maintaining a diverse set of samples in the memory bank.

%% file: table/mmlu.tex
\begin{table*}[t]
\centering
\resizebox{\linewidth}{!}{
\begin{tabular}{l r c c c c c r}
\toprule
\textbf{Models}&\textbf{Methods} & \textbf { Humanities } &\textbf  { STEM } & \textbf { Social Sciences } & \textbf  { Other } & \textbf { Average } &\textbf {Time}\\
\hline 
\multirow{4}{*}{\centering\textbf{LLaMA-2-7B}} & Zero-Shot &49.23&34.73&52.77&49.12&45.37&-\\
& Few-shot &50.54&36.80&54.07&49.76&46.75&$\times$\textbf{0.99}\\
& Self-ICL &48.61&34.86&49.98&48.94&44.64&$\times$47.28\\
& Ours &\textbf{51.76}&\textbf{37.52}&\textbf{54.12}&\textbf{50.01}&\textbf{47.33}&$\times$1.00\\
\hline
\multirow{4}{*}{\centering\textbf{Mistral-7B}} & Zero-Shot &62.93&48.29&66.97&61.58&58.83&-\\
& Few-shot &66.07&49.55&72.18&\textbf{65.11}&61.90&$\times$1.00\\
& Self-ICL &62.62&48.10&67.16&60.85&58.56&$\times$101.68\\
& Ours &\textbf{66.62}&\textbf{51.36}&\textbf{74.30}&64.11&\textbf{62.80}&\textbf{$\times$1.00}\\
\hline
\multirow{4}{*}{\centering\textbf{OpenChat-7B}} & Zero-Shot &67.50&49.72&71.67&63.97&61.90&-\\
& Few-shot &70.31&\textbf{51.78}&73.01&66.33&64.05&$\times$1.00\\
& Self-ICL &67.74&49.30&70.74&63.22&61.44&$\times$134.87\\
& Ours &\textbf{70.67}&51.40&\textbf{74.66}&\textbf{66.77}&\textbf{64.47}&\textbf{$\times$1.00}\\
\hline
    \end{tabular}
}
\caption{\label{tab: mmlu}Accuracy (\%) on the MMLU benchmark with different models and different methods. Time: the multiples of time spent by each method in reasoning the entire benchmark compared to DAIL. We omit the comparison with Zero-Shot in terms of time. The selection strategy of our reported result is DPP, and the deletion strategy is Diverse. \textbf{Bold}: the best results. We report the template in Appendix \ref{prompt} and the detailed results in Appendix \ref{detailedresults}.}
\end{table*}

%% file: paper/exp.tex
\input{table/bbh}
\section{Experiments}
\subsection{Baselines}
\noindent \textbf{Zero-Shot}\quad 
Zero-shot inference. The model directly process the query with no demonstration.

\noindent \textbf{CoT}\quad 
Chain-of-Thought \cite{wei2022chain}, an easy and effective method that utilizes specific prompts to stimulate the model's own capabilities.

\noindent \textbf{Few-Shot}\quad 
Few-shot inference. For each task, meticulously crafted demonstrations are provided by humans. Note that the comparison between Few-Shot and other baselines is not entirely fair, as Few-Shot requires additional external information.

\noindent \textbf{Self-ICL}\quad 
A zero-shot ICL method allows the model to generate new samples as demonstrations based on the query \citep{chen2023self}.

\subsection{Datasets}
\noindent \textbf{MMLU} \citep{hendrycks2020measuring} \quad 
Commonly used to evaluate the common sense reasoning ability of LLMs, MMLU consists of multiple-choice questions from various domains. 
It includes 57 subsets covering subjects in science, technology, humanities, and other areas.
Each subset has four demonstrations.

\noindent \textbf{BBH} \citep{suzgun2022challenging} \quad 
Derived from a subset of tasks within the BIG-Bench benchmark \citep{srivastava2022beyond}, BBH includes tasks where existing LLMs struggle to reach average human-rater performance.
We focus on the multiple-choice tasks, as done in \citet{chen2023self}. 
The demonstrations are provided in \citet{chen2023self}, and each subset has three demonstrations.

\subsection{Models}       
For MMLU, we utilize LLaMA-2-7B \citep{touvron2023llama}, Mistral-7B \citep{jiang2023mistral} and OpenChat-3.5-7B \citep{wang2023openchat} as our backbone.
They are the latest lightweight models with powerful capabilities.
For BBH, we employ gpt-3.5-turbo-instruct and gpt-4-1106-preview from the GPT family \citep{chatgpt}, which are currently the most popular and influential LLMs.

\subsection{Implementation details}
The number of candidates for DPP is 10.
We set $\alpha$ to 0.1 for MMLU. The size of the memory bank $M$ is 2,000.
We adopt the decoding strategy from \citet{chen2023self}.
For MMLU, we set the number of demonstrations to four. For BBH, we set the number of demonstrations to three, consistent with \citet{chen2023self}, and we use the same prompt as it.
In BBH, we cannot obtain the logits of gpt-3.5-turbo-instruct and gpt-4-1106-preview, so we omit the Entropy Score and solely leverage the Selection Score to choose the demonstrations for DAIL, which may lead to a decline in its performance.
For Sentence-BERT, we use the mostly used checkpoint from Huggingface\footnote{\href{https://huggingface.co/sentence-transformers/all-mpnet-base-v2}{https://huggingface.co/sentence-transformers/all-mpnet-base-v2}} to get the hidden states of the queries.
All the experiments are completed on NVIDIA A100-40G GPUs.

%% file: table/bbh.tex
\begin{table*}[t]
\centering
\resizebox{\textwidth}{!}{
\begin{tabular}{lccccc@{\hspace{1cm}}cccc}
\toprule
\multirow{2}{*}{\textbf{BBH Tasks}} & \multicolumn{5}{c}{\textbf{gpt-3.5-turbo-instruct}} & \multicolumn{4}{c}{\textbf{gpt-4-1106-preview}} \\
\cmidrule(lr){2-6} \cmidrule(lr){7-10} 
& \textbf{Zero-Shot} & \textbf{CoT}& \textbf{Few-Shot}& \textbf{Self-ICL}& \textbf{Ours}& \textbf{Zero-Shot} & \textbf{Few-Shot}& \textbf{Self-ICL}& \textbf{Ours} \\
\hline
Boolean Expressions &84.80&85.20&89.60&88.40&85.60&67.20&93.20&92.80&94.00\\
Causal Judgement &42.25&55.08&63.64&12.30&57.22&73.80&69.19&69.19&70.27\\
Date Understanding &59.20&44.80&52.20&57.60&55.20&48.40&73.20&74.80&79.20\\
Disambiguation QA &60.00&50.40&63.60&63.20&62.40&71.20&79.20&80.40&65.60\\
Formal Fallacies &52.00&52.40&54.80&50.40&52.00&70.40&79.60&76.00&80.00\\
Geometric Shapes &34.00&34.40&45.60&36.40&32.80&28.40&43.60&36.69&49.40\\
Hyperbaton &82.40&71.20&65.60&82.80&80.00&73.60&80.80&88.00&87.20\\
Logical Deduction (five objects) &42.00&36.40&38.00&38.40&42.00&44.80&63.20&70.40&73.60\\
Logical Deduction (seven objects) &41.60&28.40&38.80&34.80&42.40&45.20&60.00&67.20&64.40\\
Logical Deduction (three objects) &56.00&60.40&60.40&59.20&55.20&86.80&88.80&94.00&92.40\\
Movie Recommendation &74.80&77.20&78.40&76.00&71.08&80.40&92.00&80.40&92.00\\
Navigate &42.80&52.40&50.80&64.80&53.20&71.60&72.80&75.20&75.20\\
Penguins in a Table &51.37&59.59&52.74&55.48&50.68&74.66&76.03&80.82&80.14\\
Reasoning about Colored Objects &54.80&75.20&57.20&56.40&56.00&86.80&86.00&82.80&84.40\\
Ruin Names &70.80&41.20&72.40&64.80&67.34&58.63&90.80&88.00&89.20\\
Salient Translation Error Detection &41.60&46.40&51.60&51.20&45.20&68.40&67.60&68.40&69.20\\
Snarks &63.48&61.80&58.40&60.67&64.61&84.66&86.52&82.02&90.12\\
Sports Understanding &62.00&63.20&86.40&50.00&81.60&84.80&88.80&85.20&90.40\\
Temporal Sequences &20.80&36.00&38.80&32.80&40.00&97.60&100.00&99.20&100.00\\
Tracking Shuffled Objects (five objs) &18.00&24.80&17.20&16.40&21.20&36.40&33.60&28.23&35.08\\
Tracking Shuffled Objects (seven objs) &17.60&40.90&12.40&12.40&14.40&35.60&28.80&28.05&32.93\\
Tracking Shuffled Objects (three objs) &32.40&46.00&32.40&36.80&33.60&49.20&41.20&33.87&38.21\\
Web of Lies &15.20&38.80&50.00&38.40&52.00&48.00&77.20&52.40&62.80\\
\hline
All Tasks (avg)&48.69&50.08&53.21&49.55&52.86&64.07&72.50&70.81&73.47\\
\bottomrule
\end{tabular}
}
\caption{\label{tab: bbh}Accuracy (\%) on the BBH benchmark of gpt-3.5-turbo-instruct and gpt-4-1106-preview. The selection strategy of our reported result is DPP, and the deletion strategy is Diverse. The results of Zero-Shot and Self-ICL of gpt-3.5-turbo-instruct are extracted from \citet{chen2023self}. We report the cost in Appendix \ref{bbhcost}.}
\end{table*}

%% file: paper/exp_res.tex
\subsection{Results on MMLU}
Table \ref{tab: mmlu} shows the main results of each method on MMLU with different models.
We have the following observations:
\paragraph{Self-ICL performs poorly on MMLU.}
The performance of Self-ICL on MMLU is consistently inferior to that of Zero-Shot across various models.
This suggests that Self-ICL may require stronger model generative capabilities, and the models we choose may not be sufficient to generate high-quality demonstrations.
Addressing this issue might necessitate the use of larger and more powerful models, which could impose certain cost concerns on model deployment.

\paragraph{DAIL achieves State-of-the-Art (SOTA) results.}
DAIL stands out by significantly enhancing the model performance over Zero-Shot and can even surpass Few-Shot with no external information.
Furthermore, DAIL is effective across various models, indicating that it does not rely on the model's generative capabilities and possesses strong generalization ability.
\paragraph{DAIL brings no additional inference time.}
The inference speed of Self-ICL is quite slow due to the substantial amount of time required to generate demonstrations for each query.
DAIL surpasses Self-ICL in inference speed hundreds of times and is comparable to Few-Shot inference.
This remarkable efficiency brings a substantial reduction in deployment costs for real-world applications.
\paragraph{DAIL is a practical method for real-world applications.}
In scenarios with limited resources, DAIL presents a feasible solution capable of acquiring high-quality demonstrations at a minimal cost.
It consistently enhances the capabilities of ICL, making it an effective and efficient approach for real-world applications.
\subsection{Results on BBH}
Table \ref{tab: bbh} shows the main results of each method on the BBH benchmark with different models.
We have the following observations:
\paragraph{Self-ICL performs well but sometimes harms the model performance.}
Overall, compared to Zero-Shot, Self-ICL can achieve decent improvements. However, in some tasks, it can still cause substantial damage to the performance of the model (Causal Judgement, 42.25\% $\rightarrow$ 12.30\%).
Even with a powerful model, the demonstrations generated by Self-ICL may be unsatisfactory, leading to a decline in the performance of ICL.
The instability makes it impossible to play a role in real-world applications.

\paragraph{DAIL Significantly Boosts ICL Performance.}
For gpt-3.5-turbo-instruct, DAIL outperforms Zero-Shot by 4.17\% and is only 0.35\% lower than Few-Shot. For gpt-4-1106-preview, DAIL surpasses Zero-Shot by 9.4\% and exceeds Few-Shot by 0.97\%. It demonstrates the impressive capability and generalizability of DAIL.

%% file: paper/analysis.tex
\section{Analysis}
\begin{figure}[t]
\centering
\includegraphics[width=0.8\hsize]{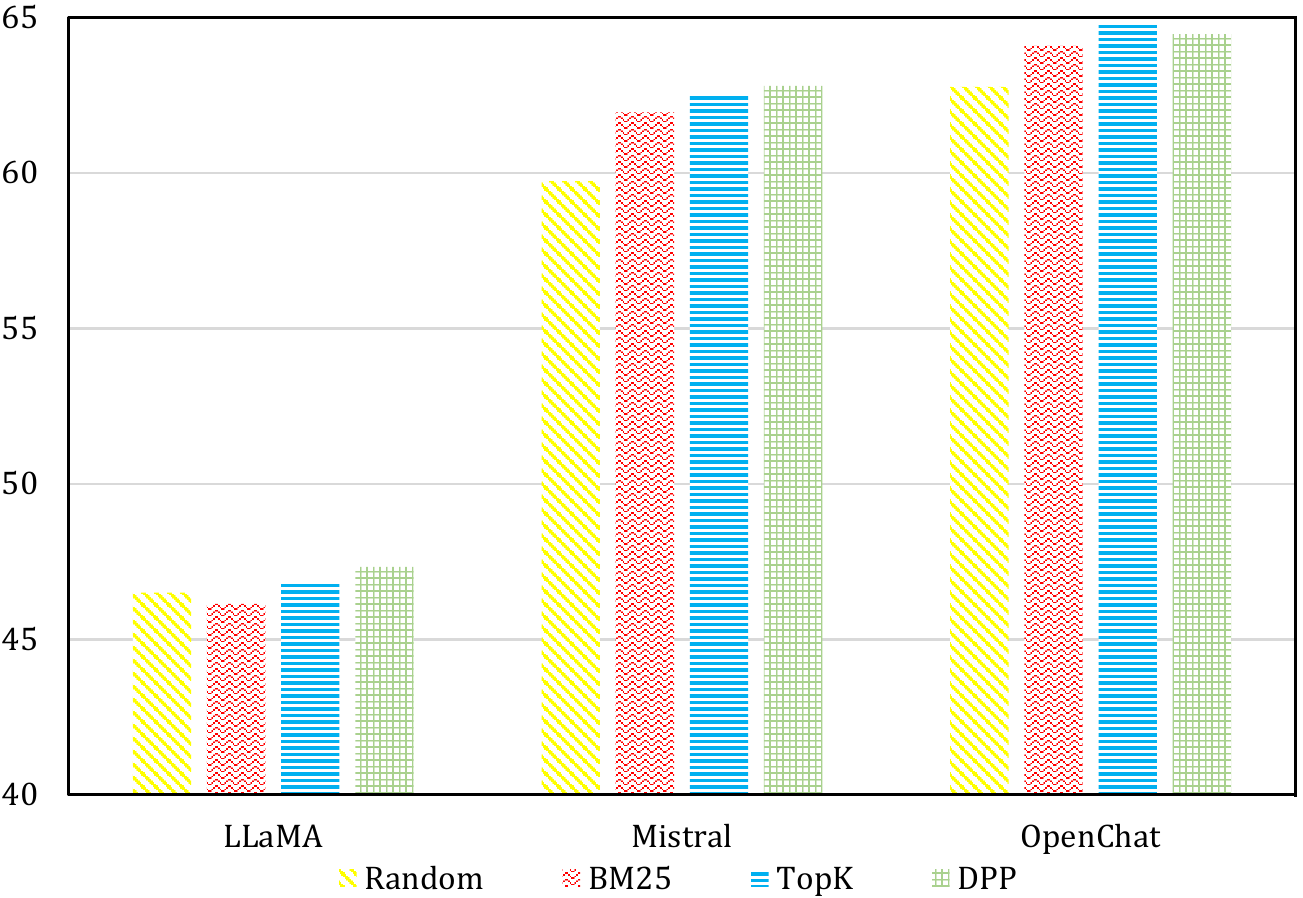}
\caption{Accuracy (\%) on MMLU with different selection strategies.}
\label{fig:select-strategy}
\end{figure}
\begin{figure}[t]
\centering
\includegraphics[width=0.8\hsize]{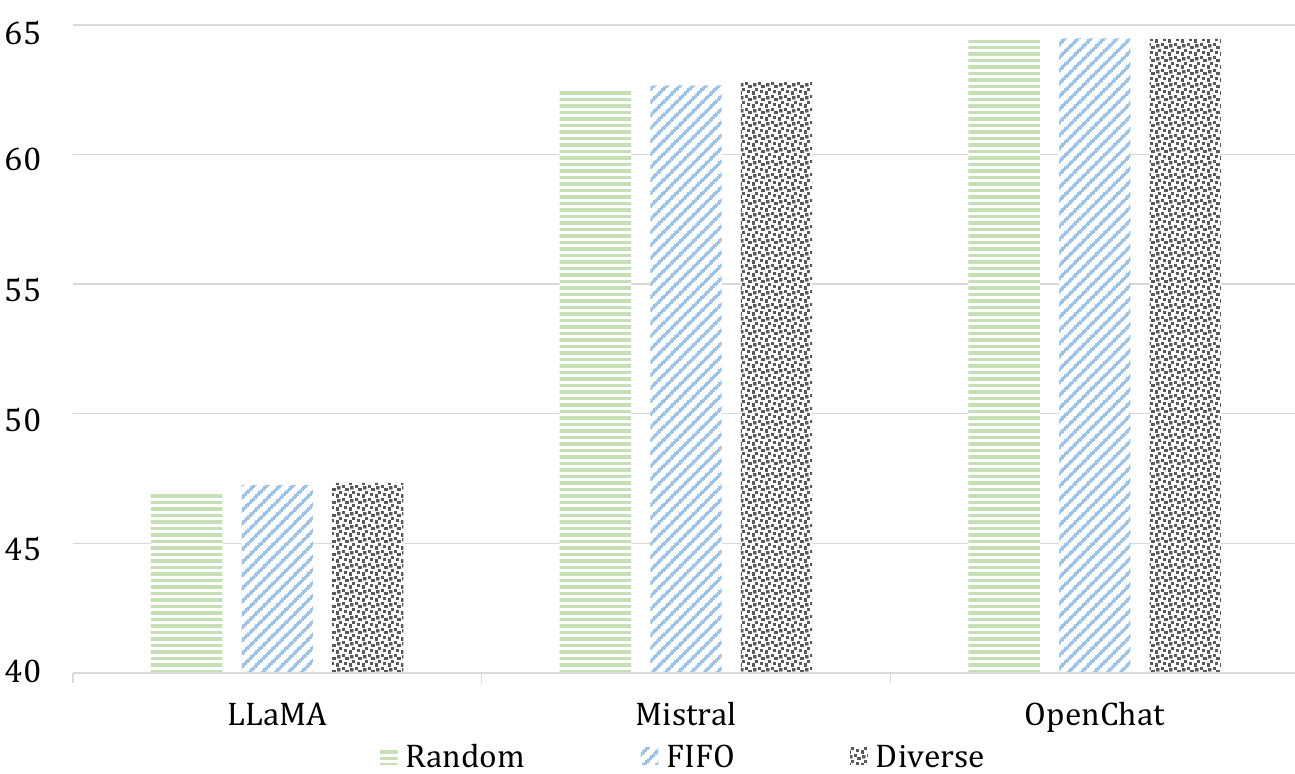}
\caption{Accuracy (\%) on MMLU with different deletion strategies.}
\label{fig:deletion-strategy}
\end{figure}
\begin{figure}[t]
\centering
\includegraphics[width=0.8\hsize]{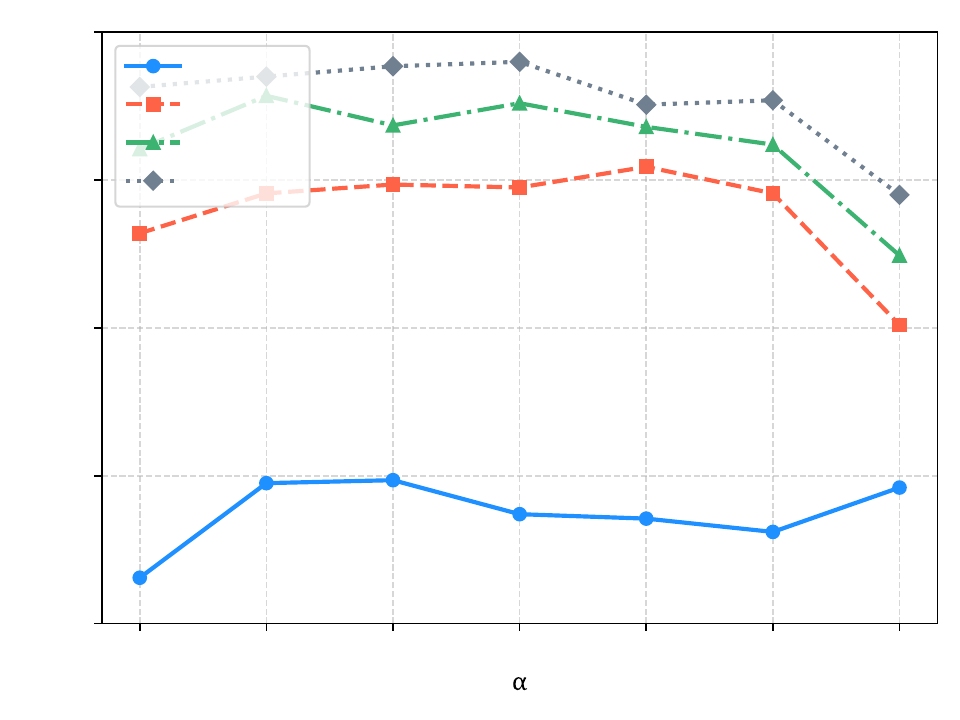}
\caption{Accuracy (\%) on MMLU with different $\alpha$. We use Mistral as the base model.}
\label{fig:alpha}
\end{figure}
\begin{figure}[t]
\centering
\includegraphics[width=0.8\hsize]{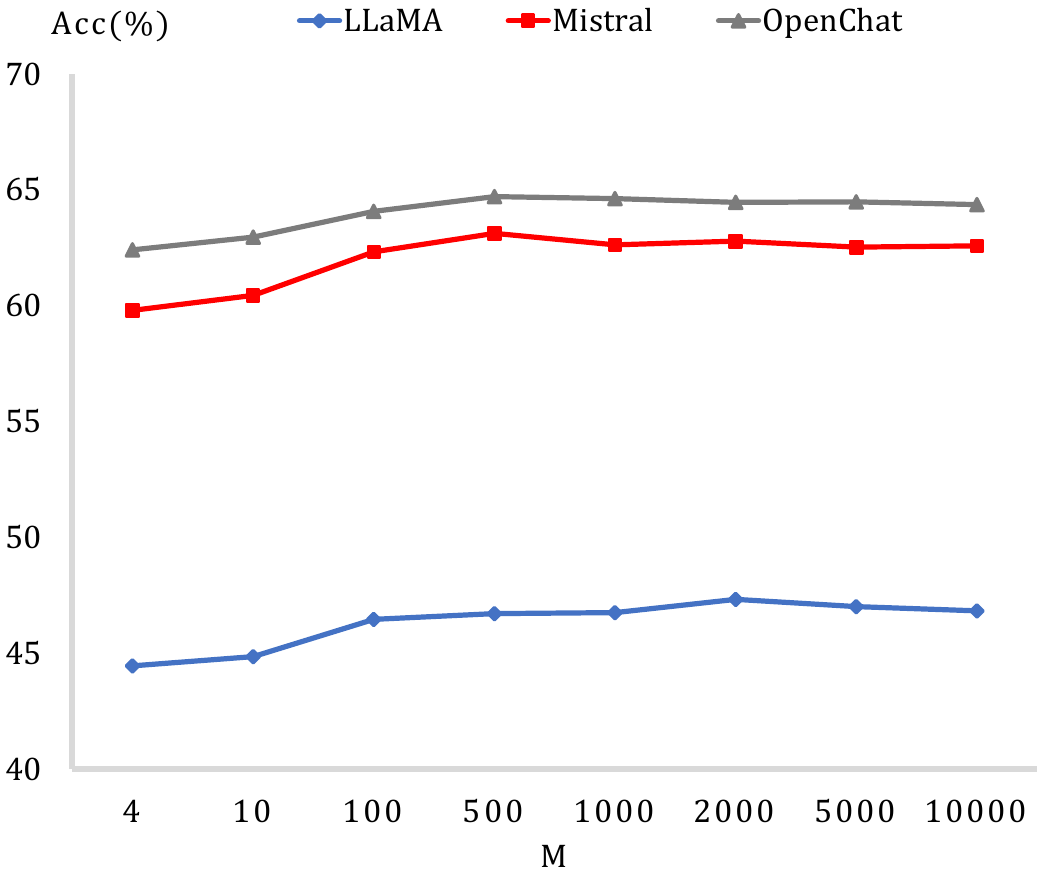}
\caption{Accuracy (\%) on MMLU with different $M$.}
\label{fig:M}
\end{figure}

\subsection{Impact of Selection Strategy}
\label{select}
The selection strategy are important for the success of DAIL.
We propose four selection strategies and we will compare the effects of them on the performance of ICL.
We conduct experiments employing different models and selection strategies on the MMLU Benchmark, Figure \ref{fig:select-strategy} shows the results of our experiments.
The results reveal that the DPP outperforms other selection strategies, with TopK being the closest competitor.
This indicates that semantic similarity plays a crucial role in DAIL.
DPP can increase the diversity of the demonstrations while maintaining the semantic similarity, which may be the primary reason for its effectiveness.
Although BM25 can compute the semantic similarity of the demonstrations and the test sample, its capability to represent semantic similarity is weaker than TopK.
Consequently, while BM25 generally outperforms Random, it does not match the performance achieved by TopK and DPP.

\subsection{Impact of Deletion Strategy}
We investigate the impact of deletion strategy using different models on the MMLU Benchmark, and the results are presented in Figure \ref{fig:deletion-strategy}.
Various deletion strategies show a minor impact on the model's performance, with the Diverse deletion strategy slightly outperforming others.
Nevertheless, we suggest that in a more dynamic environment, where we need to process various test samples from different users, the effectiveness of the Diverse deletion strategy may become more apparent.

\subsection{Impact of $\alpha$}
$\alpha$ is a crucial hyper-parameter in DAIL, balancing the Selection and Entropy scores.
To explore the impact of $\alpha$, we conduct experiments using Mistral on the MMLU Benchmark, and the results are depicted in Figure \ref{fig:alpha}.
In the case of the Random selection strategy, a notable improvement in model performance is observed as $\alpha$ increases from 0 to 0.01.
Subsequently, as $\alpha$ further increases, performance fluctuates slightly but consistently remains higher than the result when $\alpha$ is set to 0.
This reveals that incorporating the Entropy Score enhances the model's performance.
When considering other selection strategies, the model's performance tends to rise initially and then decline.
This is because when $\alpha$ is small, the combined effect of the Selection Score and the Entropy Score leads to a better selection of demonstrations.
However, when $\alpha$ is too large, the excessive weight of the Entropy Score reduces the impact of the Selection Score, and make the performance of ICL decrease.
Thus, choosing a suitable $\alpha$ is crucial, and we find that 0.1 is a good value.
\subsection{Impact of $M$}
The size of the Memory Bank ($M$) is a critical factor influencing DAIL's performance. 
Figure \ref{fig:M} displays the results of experiments with different models and $M$ on the MMLU Benchmark.
We can conclude from the figure that a too small $M$ can harm the performance of DAIL, which may be because the limited number of samples in the memory bank makes it hard to find sufficient similar and diverse demonstrations.
This limitation is alleviated as $M$ reaches 500.
Below this threshold, DAIL's performance improves with the increase of $M$, and above this threshold, the performance of DAIL stabilizes at a satisfactory level.
Considering that a $M$ that is too large will make the selection process slower and increase the cost of sample storage, we set $M$ to 2,000.

%% file: paper/related.tex
\section{Related Work}
\subsection{Understanding ICL}
In recent years, much research has delved into scaling up parameters and training data for LLMs, uncovering emergent capacities such as instruction-following, In-context Learning (ICL), and chain-of-thought.
In the realm of ICL, researchers focus on optimal demonstration selection, boosting ICL capabilities, and understanding underlying mechanisms.
As highlighted by \citet{pmlr-v139-zhao21c}, the instability of ICL is a critical problem, where factors like prompt format, demonstration examples, and example order significantly impact performance.
Despite being a challenging problem, there have been many efforts to address optimal sample selection using heuristic \citep{liu-etal-2022-makes,su2022selective} and model-based methods \citep{lu-etal-2022-fantastically,wu-etal-2023-self,levy-etal-2023-diverse}.
From another perspective, many researchers are considering enhancing the capabilities of ICL.
For example, \citet{min-etal-2022-metaicl} enhance the performance of ICL by reducing its distance from pre-training tasks, and \citet{zhao2021calibrate} eliminate the biases of some specific labels introduced by demonstrations in ICL, thereby making the distribution of label logits closer to the actual situation.

\subsection{Zero-shot ICL}
Despite the success of ICL, many studies have highlighted that the model’s performance is sensitive to the choice of demonstrations.
Although researchers have delved into the optimization of prompts and demonstrations \citep{min2022rethinking,min2022metaicl,zhao2021calibrate,chen2022improving,min2022noisy}, the reliance on a substantial amount of annotated data for demonstrations in ICL introduces additional data collection costs.
Consequently, there is a growing interest in exploring the generative capabilities of LLMs as a method to mitigate the dependency on external information.
Addressing the challenge of few-shot Chain-of-Thought (CoT) without human annotations, \citet{zhang2022automatic} leverage zero-shot CoT for demonstration construction. However, their approach still depends on an existing training set as input to zero-shot CoT. \citet{lyu2023z} explore pseudo-input generation from a raw text corpus but rely on external sources to construct pseudo-inputs. Similarly, \citet{kim2022self} investigates the generation of pseudo-inputs by the LLM itself but requires access to the label set and conditioning the language model on a label provided in the prompt.
\citet{chen2023self} propose a method that does not require any external information but still has the problem of relying on model generation capabilities and slow inference speed.

%% file: paper/conclusion.tex
\section{Conclusion}
In this paper, we analyze some previous zero-shot ICL methods and point out their shortcomings in terms of stability and inference time. To address these challenges, we propose DAIL, a simple yet effective zero-shot ICL method.
Our experiments demonstrate that DAIL can significantly enhance the performance of ICL without any external information and bring no inference latency, which indicates that DAIL has substantial potential in real-world applications.

%% file: paper/appendix.tex
\newpage
\section{Appendix}
\subsection{Analyis about the order of samples}
In our experiments, we do not shuffle the order of the samples.
According to our analysis, the order of samples is not expected to have a significant impact on the results.

Considering an ideal scenario where the memory bank is infinitely large, and we only use cos-similarity for demonstration selection. Additionally, each sample selects only one demonstration. Since cosine similarity is symmetric, if the optimal choice for sample A is sample B, then the optimal choice for sample B is sample A.
However, either sample A or sample B must enter the memory bank first, while the other enters later.
Assuming sample A enters the memory bank first, when the model infers sample A, it will select the sub-optimal sample C. If sample C is not in the memory bank, it will select an even less optimal choice. But at this point, sample B can choose its optimal solution, which is sample A, and sample C can at least choose its sub-optimal solution, which is sample A.

Following this logic, if one sample selects a better choice, then another sample will inevitably have to select a worse choice. Consequently, regardless of the order in which samples enter the memory bank, the quality of demonstrations selected by all samples will be relatively consistent overall. Therefore, the performance of the entire dataset will not vary significantly.

We conduct a series of experiments to investigate this, and the results is shown in Table \ref{tab: seq}.
\input{table/seq}

From the results, it can be observed that shuffling the order does not have much impact, except when shuffling the entire dataset. This is because that some useful samples are removed before a subset is complete, leading to a slight decrease in model performance. However, this issue can be easily resolved by increasing the memory size $M$. Increasing $M$ does not significantly increase costs, so this problem can be perfectly addressed.
\subsection{Some results on other tasks}

\subsection{Details of Experimental Cost}
\label{bbhcost}
\input{table/bbhtime}

\subsection{Prompt}
\label{prompt}
We present the prompt for the MMLU benchmark in Table \ref{tab: prompt_mmlu}.
\input{table/prompt}

\subsection{Detailed results}
\label{detailedresults}
We provide detailed results for the MMLU benchmark in Table \ref{tab: detail}.
\input{table/detail_results}

%% file: table/seq.tex
\begin{table}[h]
\centering
\resizebox{\textwidth/2}{!}{
\begin{tabular}{lccccc}
\toprule
\textbf{Methods}&\textbf{Humanities} &\textbf{STEM}&\textbf{Social Sciences}&\textbf{Other}&\textbf{Average}\\
\hline
Zero-shot&62.93&	48.29	&66.97	&61.58&	58.83\\
Samples&66.48&	51.13&	73.39&	65.11&	62.75\\
Subsets&66.58	&51.20	&72.87&	65.26	&62.75\\
All ($M$=2000)&65.79&	50.24	&73.38&	64.11	&62.07\\
All ($M$=10000)&65.79&	50.24	&73.38&	64.11	&62.74\\
Ours&66.62&	51.36&	74.30&	64.11&	62.80\\
\bottomrule
\end{tabular}
}
\caption{\label{tab: seq}Results of different orders of the sample. Samples: shuffling the order of samples within each subset. Subsets: shuffling the order in which subsets appear. All: shuffling the order of the entire dataset. $M$: the size of the memory bank. Base model: Mistral-7B.}
\end{table}

%% file: table/bbhtime.tex
\begin{table}[h]
\centering
\resizebox{\textwidth/2}{!}{
\begin{tabular}{lcccc}
\toprule
\textbf{Methods}&\textbf{Zero-Shot} &\textbf{Few-Shot}&\textbf{Self-ICL}&\textbf{Ours}\\
\hline
\multicolumn{5}{c}{\textbf{gpt-3.5-turbo-instruct}}\\
\hline
Input Tokens&763K&2,499K&4,303K&2,765K\\
Output Tokens&15K&15K&1,613K&15K\\
Cost&1.15&3.75&9.68&4.14\\
\hline
\multicolumn{5}{c}{\textbf{gpt-4-1106-preview}}\\
\hline
Input Tokens&940K&2,352K&4,992K&2,577K\\
Output Tokens&15K&15K&1,491K&15K\\
Cost&9.85&23.97&94.65&26.22\\
\bottomrule
\end{tabular}
}
\caption{\label{tab: cost}The number of consumed tokens and the cost (in US dollars) of experiments, the results of SelfICL and Zero-Shot under gpt-3.5-turbo-instruct are estimated based on the cost reported in \citet{chen2023self}.}
\end{table}

%% file: table/prompt.tex
\begin{table}[h]
\centering
\begin{tabular}{|l|}
\hline
\textbf{Demonstration:} \\
Question: A person wants to start saving money \\so that they can afford a nice vacation at the\\ end of the year. After
looking over their budget \\ and expenses, they decide the best way\\ to save money is to\\
A. make more phone calls\\B. quit eating lunch out\\C. buy less with monopoly money\\
D. have lunch with friends\\
Answer: B\\
\hline
\textbf{Test sample:} \\
Question: The complete resynthesis of phospho-\\creatine after very  high intensity exercise\\ normally takes:\\
A. about 10 seconds.\\ B. about 30 seconds. \\C. about 1 minute.\\
D. about 4 minutes.\\
Answer: \\
\hline
\end{tabular}
\caption{\label{tab: prompt_mmlu} Prompt for MMLU.}
\end{table}

%% file: table/detail_results.tex
\begin{table*}[t]
\centering
\small
\renewcommand{\arraystretch}{1.2} 
\resizebox{\textwidth}{!}{
    \begin{tabular}{l | c c c c | c c c c | c c c c}
    \toprule
\textbf{Subsets} & \multicolumn{4}{c|}{\textbf{LLaMA-2-7B}} & \multicolumn{4}{c|}{\textbf{Mistral-7B}} & \multicolumn{4}{c}{\textbf{Openchat-7B}} \\
    \cmidrule{2-5} \cmidrule{6-9} \cmidrule{10-13}
    & \multicolumn{1}{c}{\multirow{2}{*}{ZS}} & \multicolumn{1}{c}{\multirow{2}{*}{FS}} & \multicolumn{1}{c}{\multirow{2}{*}{Self}} & \multicolumn{1}{c|}{\multirow{2}{*}{DAIL}} & \multicolumn{1}{c}{\multirow{2}{*}{ZS}} & \multicolumn{1}{c}{\multirow{2}{*}{FS}} & \multicolumn{1}{c}{\multirow{2}{*}{Self}} & \multicolumn{1}{c|}{\multirow{2}{*}{DAIL}} & \multicolumn{1}{c}{\multirow{2}{*}{ZS}} & \multicolumn{1}{c}{\multirow{2}{*}{FS}} & \multicolumn{1}{c}{\multirow{2}{*}{Self}} & \multicolumn{1}{c}{\multirow{2}{*}{DAIL}} \\
    &&&&&&&&&&&& \\
    \midrule
abstract algebra&25.25&28.28&24.24&28.28&28.28&28.28&28.28&27.27&31.31&33.33&31.31&30.30\\
anatomy&43.28&43.28&41.04&44.03&57.46&59.70&59.70&61.19&61.19&64.18&55.97&63.43\\
astronomy&45.03&44.37&42.38&45.03&56.29&61.59&56.29&62.91&65.56&69.54&63.58&68.87\\
business ethics&49.49&50.51&42.42&44.44&52.53&53.54&54.55&57.58&62.63&59.60&61.62&61.62\\
clinical knowledge&53.03&50.76&51.89&54.17&68.18&71.59&66.67&70.45&70.08&68.56&72.73&71.59\\
college biology&48.25&52.45&46.85&52.45&71.33&72.73&69.93&72.73&74.13&76.22&72.03&77.62\\
college chemistry&28.28&32.32&28.28&33.33&41.41&47.47&43.43&46.46&39.39&52.53&47.47&47.47\\
college computer science&37.37&44.44&35.35&38.38&47.47&40.40&48.48&47.47&45.45&47.47&43.43&46.46\\
college mathematics&30.30&31.31&28.28&31.31&38.38&40.40&39.39&41.41&30.30&30.30&32.32&27.27\\
college medicine&40.70&40.12&44.77&41.28&58.72&63.37&58.72&61.63&62.79&65.70&63.37&64.53\\
college physics&21.78&26.73&21.78&26.73&44.55&35.64&43.56&33.66&35.64&41.58&35.64&39.60\\
computer security&48.48&53.54&60.61&59.60&71.72&73.74&70.71&78.79&70.71&74.75&66.67&74.75\\
conceptual physics&35.04&39.32&41.88&38.89&46.58&54.27&47.01&54.70&56.41&55.13&57.26&55.98\\
econometrics&30.09&34.51&28.32&32.74&38.05&46.90&36.28&46.02&46.90&46.90&46.02&53.98\\
electrical engineering&40.28&42.36&43.75&45.83&52.78&55.56&54.17&56.25&50.69&50.69&52.08&52.08\\
elementary mathematics&26.79&28.65&27.59&26.53&36.60&36.34&36.34&36.87&42.97&41.11&42.18&44.03\\
formal logic&24.00&24.80&21.60&25.60&38.40&33.60&40.80&38.40&39.20&43.20&41.60&44.00\\
global facts&35.35&39.39&41.41&40.40&33.33&37.37&34.34&32.32&27.27&30.30&31.31&31.31\\
high school biology&50.16&54.69&50.49&53.07&69.26&76.38&68.61&77.02&75.40&81.55&76.05&79.29\\
high school chemistry&35.15&35.64&33.17&33.17&45.54&46.53&43.07&49.50&44.55&46.04&45.54&49.01\\
high school computer science&39.39&36.36&36.36&45.45&64.65&61.62&62.63&63.64&66.67&70.71&62.63&69.70\\
high school european history&57.32&55.49&54.88&54.88&70.73&73.17&71.34&75.00&79.88&78.05&78.05&80.49\\
high school geography&60.41&60.91&61.42&64.47&73.10&78.68&73.60&83.25&75.63&79.70&76.14&82.74\\
high school government and politics&65.62&65.10&61.46&60.94&83.33&88.54&82.29&88.02&87.50&89.06&84.38&89.06\\
high school macroeconomics&40.87&41.65&40.87&44.47&55.53&64.01&56.30&67.61&62.72&62.98&62.21&66.32\\
high school mathematics&25.65&22.68&28.62&25.65&30.48&28.62&30.48&28.62&30.48&28.25&30.86&28.25\\
high school microeconomics&43.46&42.62&37.97&44.30&60.34&67.09&59.92&70.04&67.51&67.51&63.71&69.62\\
high school physics&26.67&26.00&23.33&28.00&37.33&37.33&38.67&41.33&38.67&39.33&36.00&40.67\\
high school psychology&60.29&65.07&59.56&65.07&75.92&80.51&76.10&83.27&83.27&83.82&80.33&83.27\\
high school statistics&26.98&31.63&27.44&30.23&42.33&56.28&44.19&53.49&46.98&48.37&43.72&49.77\\
high school us history&59.11&61.58&55.67&65.52&78.33&77.34&75.37&80.30&76.85&81.77&78.33&81.77\\
high school world history&59.75&60.59&61.44&63.98&75.00&75.42&69.92&77.97&81.78&80.08&80.51&80.51\\
human aging&50.45&54.50&55.86&54.50&71.17&70.27&70.27&70.72&66.22&68.92&68.02&71.17\\
human sexuality&57.69&53.08&49.23&51.54&71.54&75.38&70.00&77.69&76.15&77.69&76.15&78.46\\
international law&55.00&61.67&55.83&61.67&69.17&79.17&71.67&79.17&78.33&80.00&76.67&79.17\\
jurisprudence&54.21&54.21&51.40&57.94&69.16&72.90&67.29&74.77&75.70&79.44&71.96&75.70\\
logical fallacies&53.70&54.32&52.47&48.77&71.60&74.07&70.99&71.60&72.84&74.07&72.22&75.93\\
machine learning&34.23&31.53&27.03&33.33&44.14&38.74&40.54&52.25&49.55&45.05&48.65&44.14\\
management&61.76&62.75&63.73&64.71&66.67&77.45&66.67&75.49&79.41&84.31&77.45&85.29\\
marketing&69.96&71.67&66.52&75.11&84.55&87.55&85.41&87.12&87.12&88.84&84.55&87.55\\
medical genetics&45.45&51.52&52.53&47.47&64.65&72.73&65.66&65.66&65.66&76.77&65.66&73.74\\
miscellaneous&64.71&62.02&63.43&65.35&76.21&80.43&75.83&81.71&81.07&81.33&80.05&80.95\\
moral disputes&46.38&51.59&46.96&55.94&66.38&68.41&66.38&71.59&71.30&71.30&71.88&74.78\\
moral scenarios&24.16&21.36&24.83&24.50&24.38&37.47&24.83&24.38&36.69&47.20&43.18&46.87\\
nutrition&51.80&54.10&48.52&51.80&69.51&73.77&59.34&74.43&70.49&73.77&70.16&74.75\\
philosophy&50.00&54.19&54.84&56.45&64.84&70.97&63.87&71.94&65.81&72.26&67.74&72.26\\
prehistory&55.42&54.18&54.49&54.49&68.73&72.45&69.35&69.97&70.90&74.61&71.21&74.92\\
professional accounting&35.94&37.01&34.52&35.59&47.69&46.26&46.62&47.69&44.48&45.91&41.28&49.82\\
professional law&35.68&34.77&35.75&34.96&43.18&43.31&43.38&45.08&45.92&47.95&46.05&48.14\\
professional medicine&37.27&36.53&34.32&34.69&62.36&64.21&60.89&61.25&68.63&68.27&65.68&69.37\\
professional psychology&43.70&44.68&42.06&47.30&60.56&65.96&61.21&68.74&63.18&63.67&63.18&67.92\\
public relations&51.38&49.54&47.71&55.96&57.80&64.22&61.47&67.89&62.39&59.63&61.47&62.39\\
security studies&45.08&54.92&38.93&43.85&62.70&65.98&63.93&71.72&68.44&71.31&68.44&71.72\\
sociology&66.00&62.00&63.50&65.00&82.00&83.00&82.00&82.50&80.50&85.00&83.00&83.50\\
us foreign policy&68.69&74.75&68.69&73.74&82.83&85.86&82.83&84.85&85.86&88.89&83.84&86.87\\
virology&48.48&42.42&44.24&46.67&49.09&53.33&47.27&50.30&48.48&52.12&47.27&49.70\\
world religions&65.29&68.24&61.76&68.24&78.24&80.59&78.82&85.88&82.35&84.12&81.18&84.12\\
\midrule
Average&45.37&46.75&44.64&47.33&58.83&61.90&58.56&62.80&61.90&64.05&61.44&64.47\\
    \bottomrule
    \end{tabular}
    }
\caption{\label{tab: detail} Detailed Results (Accuracy\%) on MMLU. ZS: Zero-Shot, FS: Few-Shot, Self: Self-ICL.}
\end{table*}